# ESTIMATING SECTORAL POLLUTION LOAD IN LAGOS, NIGERIA USING DATA MINING TECHNIQUES

[1]Adesesan .B ADEYEMO, [2]Adebola.A OKETOLA, [3]Emmanuel.O ADETULA, [4]O.OSIBANJO

[1]University of Ibadan, Computer Science Department, Nigeria
Sesan_adeyemo@yahoo.com

[2]University of Ibadan, Department of Chemistry, Nigeria
bolaoketola@yahoo.com

[3]University of Ibadan, Computer Science Department, Nigeria
olumuyiwaadetula@gmail.com

[4]University of Ibadan, Department of Chemistry, Nigeria
oosibanjo@yahoo.com

**ABSTRACT**
Industrial pollution is often considered to be one of the prime factors contributing to air, water and soil pollution. Sectoral pollution loads (ton/yr) into different media (i.e. air, water and land) in Lagos were estimated using Industrial Pollution Projected System (IPPS). These were further studied using Artificial neural Networks (ANNs), a data mining technique that has the ability of detecting and describing patterns in large data sets with variables that are non- linearly related. Time Lagged Recurrent Network (TLRN) appeared as the best Neural Network model among all the neural networks considered which includes Multilayer Perceptron (MLP) Network, Generalized Feed Forward Neural Network (GFNN), Radial Basis Function (RBF) Network and Recurrent Network (RN). TLRN modelled the data-sets better than the others in terms of the mean average error (MAE) (0.14), time (39 s) and linear correlation coefficient (0.84). The results showed that Artificial Neural Networks (ANNs) technique (i.e., Time Lagged Recurrent Network) is also applicable and effective in environmental assessment study.

## 1. INTRODUCTION

Industrial pollution is one of the leading causes of pollution worldwide. Industrial pollution is a serious problem for the entire planet especially in nations which are rapidly industrializing.

There are a number of forms of industrial pollution; one of the most common is water pollution caused by dumping of industrial waste into water ways or improper containment of waste, which causes leakage into groundwater and waterways. Industrial pollution can also impact air quality, and it can enter the soil, causing widespread environmental problems. Industrial pollution hurts the environment in a range of ways, and it has a negative impact on human lives and health. Pollutants can kill animals and plants, imbalance ecosystems, degrade air quality radically, damage buildings and generally degrade quality of life. Although, pollution is an expensive, undesirable, inevitable and necessary part of human life. Even in primitive cultures, accumulated human excretory products and smoke from cooking fires causes pollution (Chaloulakou, 2007). The main factors responsible for pollution and other types of environmental deterioration are the combined effect of population, effluent and technology on which industrial development is predicated, all things being equal.

*Industry* can be defined as the manufacture or production of goods or any business activity while industrialization is the process of making a place industrially developed (Heintz, 2009). Industrialization, though desired by all the nations of the world is always at the other end opposing the environment. The era of industrial revolution was greatly welcomed and ushered in with great admiration but no sooner had man begun to enjoy the dividend of industrialization that he began to see its toll on his immediate environment, health and well being (World Bank, 1991). Man has witnessed and still witnessing many great negative impacts of industrialization and urbanization either directly or indirectly. The worrisome aspect of pollution arising from industrialization is that it is not localized i.e. pollution generated in a given environment has the tendency to migrate through the atmosphere, water and man to other parts that do not have the culprit industries, though the effect is felt much more where it emanates than elsewhere (Patricio,2000).

The fields of Ecology and Environmental Chemistry have mainly employed the use of statistical techniques for the analysis of the relationship between an observed response and set of predictors in a data set. The statistical techniques used are parametric; this implies that it requires the user to specify the predictor variables to include in the analyses. This approach to data analysis is appropriate for both parameter estimation and hypothesis testing as long as the analyst has sufficient prior knowledge to specify an appropriate parametric model (Bhargavi, 2009). However, this type of flexibility may be insufficient to allow ecologists and environmentalist to extract biological insights, generate patterns, new discovery, and relationship among variables from the data set if prior knowledge is minimal and hypotheses are not clearly developed (Bauer, 1999). Under these circumstances, exploratory analyses (i.e., analyses useful for generating hypotheses) are more appropriate than the confirmatory analyses (i.e., analyses designed to test hypotheses or estimate model parameters) typically carried out by ecologist in their research evaluation (Hastie et al. 2001).

### 1.1 POLLUTION LOAD AND POLLUTION INTENSITY

The total amount of a pollutant or combination of pollutants released into the environment (directly or indirectly through the municipal sewers or through the municipal wastes collectors and treatment network) by an industry or a group of industries in a given area during a certain period of time (WHO, 1992).

Pollution intensity is defined as the level of Pollution discharge or emissions per unit of manufacturing activity (Pandey, 2005). In calculating the Pollution Intensity, the choice of the variable to measure the level or size of manufacturing activity is very important; the IPPS database provides estimates for three alternatives measures of the level of manufacturing activities, viz; value of output, value added and employment.

Hettige et al. (1994) has shown that in the case of the US, the ranking of industrial Sectors by their pollution load is almost identical irrespective of whether the value of output or employment is used as the unit of measurement the same trend was obtainable in Lagos, Nigeria (Oketola and Osibanjo, 2007; Oketola and Osibanjo, 2009 ). Total value of output was, however judged superior to value added because the energy and materials inputs are critical in the determination of industrial pollution (Pandey, 2005).

### 1.2 DATA MINING

Data mining is the process of applying techniques such as association, clustering, classification and prediction to data with the intention of uncovering hidden patterns. This high level of relevance is viewed with data mining's ability in connecting database, artificial intelligence, and statistics together with other fields (e.g. biological sciences, physical sciences and the social sciences) (Zaiane, 1999). In contrast to standard statistical methods, data mining techniques search for interesting information without demanding a priori hypotheses. The kinds of patterns that can be discovered depend upon the data mining tasks employed. By and large, there are two types of data mining tasks: descriptive data mining that describes the general properties of the existing data; and predictive data mining that attempt to do predictions based on inference on the available data. This techniques are often more powerful, flexible, and efficient for exploratory analysis than the statistical techniques (Bregman, 2006).

### 1.3 ARTIFICIAL NEURAL NETWORKS

Artificial Neural Networks (ANNs) are massively parallel distributed processor made up of simple processing units, which has a natural propensity for storing experimental knowledge and making it available for use (Anders, 2005). An artificial neuron is an information processing unit that is fundamental to the operation of a neural network. There are three basic elements of a neuron model, which are: (i) a set of synapses connecting links, each of which is characterized by a weight or strength of its own; (ii) an adder for summing the input signals weighted by the respective synapses of the neuron; and (iii) an activation function for limiting the amplitude of the output of a neuron. A typical input-output relation can be expressed as shown in Equation 1 while Figure 1, shows the basic elements of neuron model with the help of a perceptron model.

$$net_j = \sum_{j=1}^{n} w_{ij} x_i + b_j$$

$$o_i = f_i(net_i) \qquad (1)$$

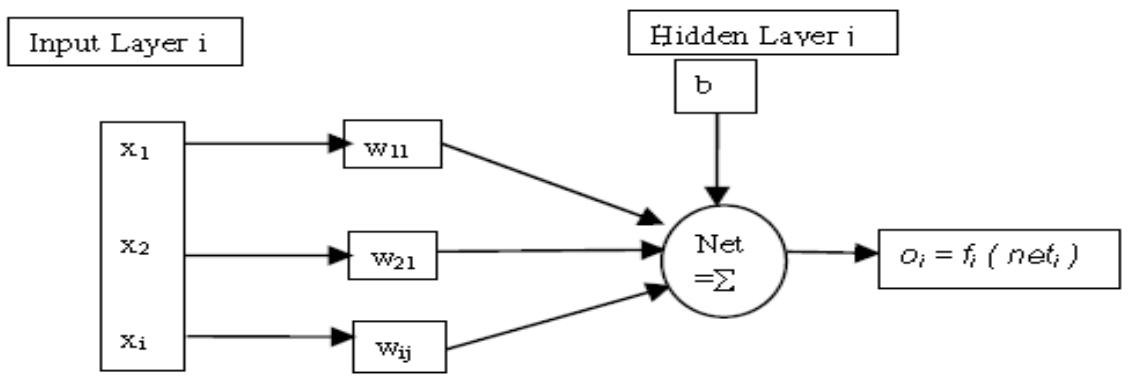

Figure 1: Model of a perceptron

where, $X_i$ = inputs to $i^{th}$ node in input, $W_{ij}$ = weight between $i^{th}$ input node and $j^{th}$ hidden node, b =bias at $j^{th}$ node, net = adder, f = activation function.

The type of transfer or activation function affects size of step taken in weight space (Martin, 2002). Use of Sigmoidal transfer function in hidden layer and linear transfer function in output layer is advantageous for extrapolation beyond range of training data (Sherod, 2003). ANNs architecture requires determination of the number of connection weights and the way information flows through the network. These are done by choosing the number of layers, number of nodes in each layer and their connectivity. A number of output nodes are fixed by the quantities to be estimated. The number of input nodes is dependent on problem under consideration and the modeller's discretion to utilize domain knowledge. The number of neurons in hidden layer is increased gradually and the performance of the network in the form of an error is monitored. It is observed that error goes on reducing as the hidden neurons are increased up to a certain limit beyond which network performance goes down in validation (Roberts, 2007).

According to Kolimogorov theorem, any continuous function with 'n' inputs and 'm' outputs can be represented exactly with three layer of ANNs containing '2n+1' nodes in the hidden layer (Minsky,1969). Another criterion for number of nodes in hidden layer is given by Lallahem as shown in Equation 2;

$$(A + 1) B + (B + 1) C \leq (1/10) D \qquad (2)$$

Where, A is the number of nodes in input layer; B is the number of nodes in hidden layer; C is the number of nodes in output layer; D is the number of training patterns.

***The aim of the study*** is therefore to develop a model using Artificial Neural Networks (ANNs), a data mining tool to investigate the sectoral pollution load in Lagos based on Industrial Pollution Projection System (IPPS) estimation; and also to assess and predict the applicability of ANNs to environmental studies.

## 2. MATERIAL AND METHODS

### 2.1 THE DESCRIPTION OF THE STUDY AREA

Lagos state is located within the low-lying coastal zone, which extends along 180 km of marine shoreline and inland to a distance of about 32km, out of which about 17% are made up of lagoons, creeks and coastal river estuaries (Adefuye et al., 2002; Onyekwelu et al., 2003). The state lies approximately between longitudes 20 42 / E to 30 42/ E and latitudes 60 22/ N to 60 42/ N (Akinsanya, 2003; Onyekwelu et al., 2003). Although, it is one of the smallest states in Nigeria, being only 0.4 % of the total land area, it is the most industrialized and the commercial capital of Nigeria. The metropolitan area covers are Lagos Island, Lagos Mainland, Mushin East (Shomolu), Mushin West, Ikeja, Badagry, Ikorodu, Agege, Alausa, Isheri and Ketu (majidun). Figure 2 shows the map of Lagos. Lagos state has about 7,000 medium and large scale industrial establishments which are scattered within the fourteen developed industrial estates in the state. The developed industrial estates as of 2000 A.D. (Arikawe-Akintola, 1987; Akinsanya, 2003) are Ikeja, Agidingbi, Amuwo Odofin (industrial), Apapa, Gbagada, Iganmu, Ijora, Ilupeju, Matori, Ogba, Oregun, Oshodi/Isolo/Ilasamaja, Surulere (light industry) and Yaba as shown in Figure 3. Lagos state been the most industrialized state in Nigeria could be inferred to be the most polluted state in terms of the level of industrialization and its living population which is over 140 million (Stock, 2008). Available data (sectoral employment figure, total value of output, pollution load and pollution intensities by medium) were obtained from Oketola and Osibanjo, (2007, 2009).

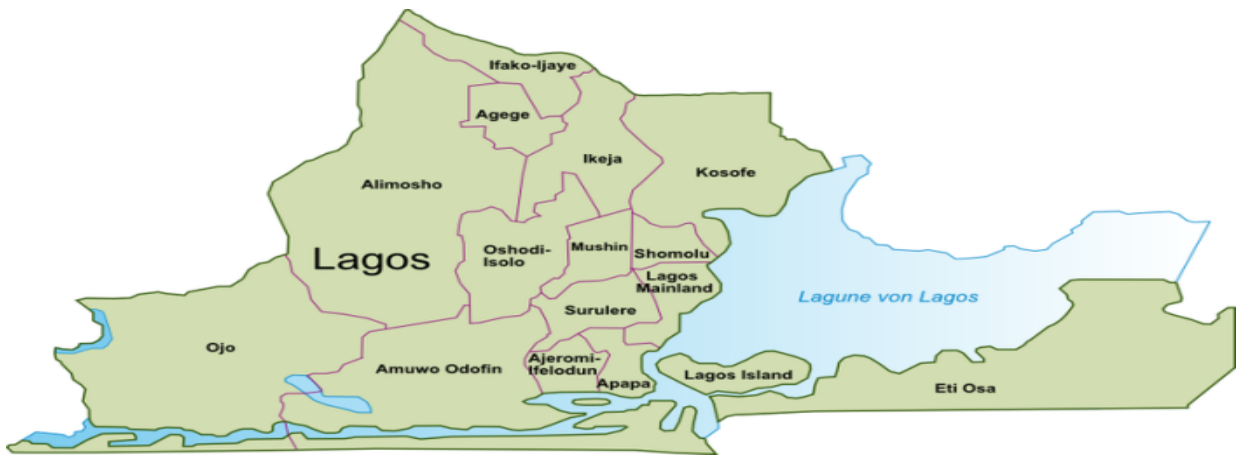

Figure 2: Map of Lagos

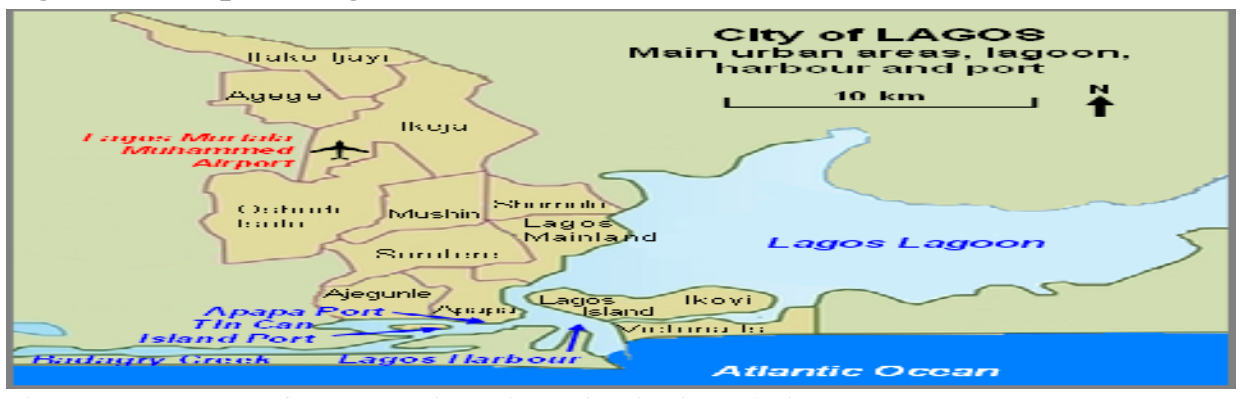

Figure 3: Map of Lagos showing the industrial estates

### 2.2 DATA ACQUISITION

The industries in Nigeria are grouped into ten major sectors by the Manufacturers Association of Nigeria (M.A.N). The grouping are Chemical and Pharmaceutical (CPH), Food, Beverage and Tobacco (FBT), Non metallic Mineral Products (NMP), Pulp and Paper Products (PPP), Basic Metal (BML), Textile, Wearing Apparel (TWA), Domestic and Industrial Products (DIP), Electrical and Electronics (EES), Wood and Wood Products (WWP), Motor Vehicle and Miscellaneous (MVM) These sectors have different sub sectors, which are based on similar production activities (Oketola, 2007). Estimated IPPS sectoral pollution loads with respect to employment and total value of output by medium were considered in this study (Oketola and Osibanjo, 2007). The pollutants are Criteria air pollutants which include sulphhur dioxide ($SO_2$), nitrogen dioxide ($NO_2$), carbon monoxide (CO), volatile organic compounds (VOC), total suspended particulate (TSP), fine particulate (FP) Water pollutants; biological oxygen demand (BOD) and suspended solids (SS) Toxic pollutants; toxic chemicals and bio-accumulative metals.

### 2.3 MODEL BUILDING

The data set in this study consists of 39 sets of variables which were made up of ten (10) categorical and twenty nine (29) continuous variables as shown in Table 1. In each model, the input data sets were *year, sector name, pollution intensity (PI), employment value for each year (1997-2005), and production output* while the output variable used for each network model was the sectoral *pollution load* with respect to employment for each pollutant.

### 2.4 THE DATA SET FORMATTING

The Artificial Neural Networks (ANNs) models were trained using the variation of the data set in order to obtain the best network which gave the expected output. The data sets were divided into three, namely: Training set, Cross Validation and Testing set.

#### 2.4.1 TRAINING DATA SET

The training data set was used to enable the network learn relationships between the input data and the expected output. It enables the system to observe, learn and develop a relationship between the input and output variables. 60% of the whole data were used for the training process. The contributions of the input variables to the output variables were also determined. It enabled us to monitor and thereby extract the cause and effect relationship between the inputs and the output of the network; and to know which input variables was the most significant and which was not. This generated insights which were used to prune the network, helping to reduce the complexity of the network and its training time. Also, during the network training phase, the system adjusted its connection /weight strengths in favour of the inputs that were the most effective in determining the output. Training set sample for the model is shown in Table 2.

#### 2.4.2 CROSS VALIDATION DATA SET

The cross-validation data set was used to periodically check how far the network has been able to learn the relationships between the inputs and the output. The use of cross validation data set in this study is an important guide against overtraining and over fitting the network. If overtraining does occur, the network loses its ability to generalize its outcomes. This is used to connote the ability of a neural network to perform very well on data sets it has not been trained with. Generalization is affected by three factors which are the number of the training set, the number of parameters of the model (number of weights) and the complexity of the problem at hand. Over fit is said to occur when a model correctly handles the training data but fails to generalize. One method that is said to estimate the performance of a model is to estimate the generalization errors of the model. If the generalization errors are of acceptable minimum value then the model generalizes well (Adefowoju, 2003).

Table 1: Variable format for the models

| S/NO | NAME OF VARIABLE | TYPE OF VARIABLE | INPUT / OUTPUT VARIABLE |
|---|---|---|---|
| 1 | Food, beverages and tobacco sector (FBT) | Categorical | Input |
| 2 | Textile, weaving apparel sector (TWA) | Categorical | Input |
| 3 | Wood and wood products sector (WWP) | Categorical | Input |
| 4 | Paper and paper products sector (PPP) | Categorical | Input |
| 5 | Chemical and pharmaceutical sector (CPH) | Categorical | Input |
| 6 | Non-metallic mineral product sector (MMP) | Categorical | Input |
| 7 | Domestic and industrial plastics sector (DIP) | Categorical | Input |
| 8 | Electrical and electronics sector (EES) | Categorical | Input |
| 9 | Basic metal sector (BM) | Categorical | Input |
| 10 | Motor Vehicles and miscellaneous sector (MVS) | Categorical | Input |
| 11 | Employment value | Continuous | Input |
| 12 | Production output | Continuous | Input |
| 13 | Pollution intensity for $SO_2$ Employment | Continuous | Input |
| 14 | Pollution intensity for $NO_2$ Employment | Continuous | Input |
| 15 | Pollution intensity for CO Employment | Continuous | Input |
| 16 | Pollution intensity for VOC Employment | Continuous | Input |
| 17 | Pollution intensity for FP Employment | Continuous | Input |
| 18 | Pollution intensity for TSP Employment | Continuous | Input |
| 19 | Pollution intensity for TCAIR Employment | Continuous | Input |
| 20 | Pollution intensity for TCLAND Employment | Continuous | Input |
| 21 | Pollution intensity for TCWATER Employment | Continuous | Input |
| 22 | Pollution intensity for TMAIR Employment | Continuous | Input |
| 23 | Pollution intensity for TMWATER Employment | Continuous | Input |
| 24 | Pollution intensity for TMLAND Employment | Continuous | Input |
| 25 | Pollution intensity for BOD Employment | Continuous | Input |
| 26 | Pollution intensity for TSS Employment | Continuous | Output |
| 27 | Pollution load for $SO_2$ Output | Continuous | Output |
| 28 | Pollution load for $NO_2$ Output | Continuous | Output |
| 29 | Pollution load for CO Output | Continuous | Output |
| 30 | Pollution load for VOC Output | Continuous | Output |
| 31 | Pollution load for FP Output | Continuous | Output |
| 32 | Pollution load for TSP Output | Continuous | Output |
| 33 | Pollution load for TCAIR Output | Continuous | Output |
| 34 | Pollution load for TCLAND Output | Continuous | Output |
| 35 | Pollution load for TCWATER Output | Continuous | Output |
| 36 | Pollution load for TMAIR Output | Continuous | Output |
| 37 | Pollution load for TMWATER Output | Continuous | Output |
| 38 | Pollution load for BOD Output | Continuous | Output |
| 39 | Pollution load for TSS Output | Continuous | Output |

Table 2: Training Data Set Sample

| SECTORS | YEARS | EMPLOYMENT | PI $SO_2$ | PI $NO_2$ | PI CO | PI VOC | PI FP | PI TSP | PI TCAIR | PI TCLAND |
|---|---|---|---|---|---|---|---|---|---|---|
| BM | 1999 | 2000 | 11363715 | 2403689 | 9001620 | 2214652 | 894782 | 10518 | 1711757 | 3070992 |
| WWP | 1997 | 457 | 324752 | 396905 | 347260 | 1039161 | 136324 | 8446 | 218457 | 25617 |
| BM | 2005 | 65000 | 11363715 | 2403689 | 9001620 | 2214652 | 894782 | 10518 | 1711757 | 3070992 |
| WWP | 1998 | 107 | 324752 | 396905 | 347260 | 1039161 | 136324 | 8446 | 218457 | 25617 |
| DIP | 1998 | 2360 | 13680464 | 3766768 | 551943 | 3318529 | 11101 | 1231 | 1867319 | 1397019 |
| PPP | 2005 | 35000 | 6344752 | 34287663 | 7066002 | 1155022 | 349944 | 5084 | 1210727 | 483003 |
| EES | 1997 | 599 | 515941 | 173935 | 307709 | 317396 | 1760 | 364 | 1711757 | 3070992 |
| TWA | 1999 | 4240 | 309169 | 496163 | 150850 | 305642 | 20120 | 3157 | 1999955 | 2383466 |
| NMP | 2003 | 66700 | 27354929 | 15576253 | 2382725 | 497374 | 23003045 | 92290 | 192218 | 200535 |

### 2.4.3 TEST DATA SET

The test data set helped to know and verify the networks performance and the effectiveness of the connection strengths established during the network training phase. The test data set basically consist of input data and a known output. This further gave opportunity for comparison and verification between the networks result and the expected result (Adefowoju, 2003).

## 2.5 NETWORK LAYERS AND PROCESSING ELEMENT (PE)

The choice of the neural network topology used in modelling the system was another point to resolve. The neural network topology describes the arrangement and structure of the neural network. The choice of topology used was a difficult decision to make. An understanding of the topology and the type of data set as a whole are very paramount. The number of hidden layers, number of epoch, momentum and learning rates were taking into consideration in this study. In considering the number of hidden layers, number of input variables and expected number of outputs in the network and the data set were split into different units in other to see how better it would train on a smaller unit. Having smaller number of hidden layers in a neural network model lowers the processing capability of the network. Also, a large number of hidden layers on a network will progressively slow down the training time. Examples indicated that training times grow exponentially with the number of dimension of the networks input for each network models. To further enhance building the model, two approaches were considered on the selection of the network size. First with a small network and then increase the hidden layers and keep increasing the hidden layers. Secondly, a more complex network would train better by setting smaller hidden layers and then increasing it to see the change and trend within the models. Both methods were considered in this study since the data set had to be considered based on sectors and pollutants.

## 2.6 LEARNING PARAMETERS

The control of the learning parameters is an unresolved problem in artificial neural networks (ANNs) research as well as in optimization theory (i.e., time involved). The goal was to reach the optimum performance in small training time. If the learning parameter rate is increased, this will not only cause a corresponding decrease in the training time, but will also increase the possibility of creating divergent iterative process and the optimum solution is not obtained (i.e.the network is memorizing ). Therefore, there is need to seek a way to find the largest possible step size that guarantee convergence. The learning parameters of the chosen network topology that fit into this study were considered to determine the best parameters setting. The conventional approach which was employed was the selection of the learning rate and a momentum term. Momentum learning is an enhancement over the straight gradient descent search by imposing a memory factor on the adaptation. This has the advantage of fast adaptation, at the same time reducing the probability of getting stuck at the local minimal. Thus the learning equation, equation 3 is

$$\Delta W_{ij}(k) = \gamma \Delta W_{ij}(K-1) - \mu \delta E(k) \qquad (3)$$

Where, $\mu$ is the learning rate and $\gamma$ is a constant (normally set between 0.5 and 0.9).

In this study, different learning rate ranging between 0.001 and 1 with the momentum term of 0.9 was used for the hidden and the output layers. Another option is whether to assign initial weights with random values or using the same weight values. Although, the choice for good initial weights for training has received very little attention as a result of the complexity involved (Ajakaye et al., 2006). To avoid symmetry conditions (normally associated with the latter approach) that can trap the search algorithm, the initial weights were started at random values. The use of varied random starting weights on each run can generate different outcome. Therefore, five independent runs were made on each topological model in order to get the best combination. On the other hand is the weight update timing which can either be done when all the exemplars of the training set have been presented (batch learning) or at each iteration (real time). The first method smoothing the gradient may offer faster learning when noise is present in the data. However, it may also average the gradient to zero and stall learning. Modification of the weight for each iteration with a small learning rate may be preferable most of the time. This approach was used during the training section.

## 2.7 PERFORMANCE MEASURES

The Neurosolutions6 software has some basic standard parameters that were used to evaluate each model performance. The performance measures used for the study are:

### 2.7.1 LEARNING CURVE

The learning curve shows how the mean square error evolves with the training iteration. It is a quantity that can be used to check the progress of learning. The difficulty of the task and how to control the learning parameters can be determined from the learning curve. When the learning curve is flat, the step size is increased to speed up learning and when the learning curve oscillates up and down, the step size is decreased. In the extreme, the error curve increase steadily upwards, showing that the learning is unstable. At this point, the network is reset. When the learning curve stabilizes after it has been set again, it could be certified to be training the datasets. If the curve looks like the sample graph shown in Figure 4, the network is said to have been well trained, while Figure 5 has not trained well.

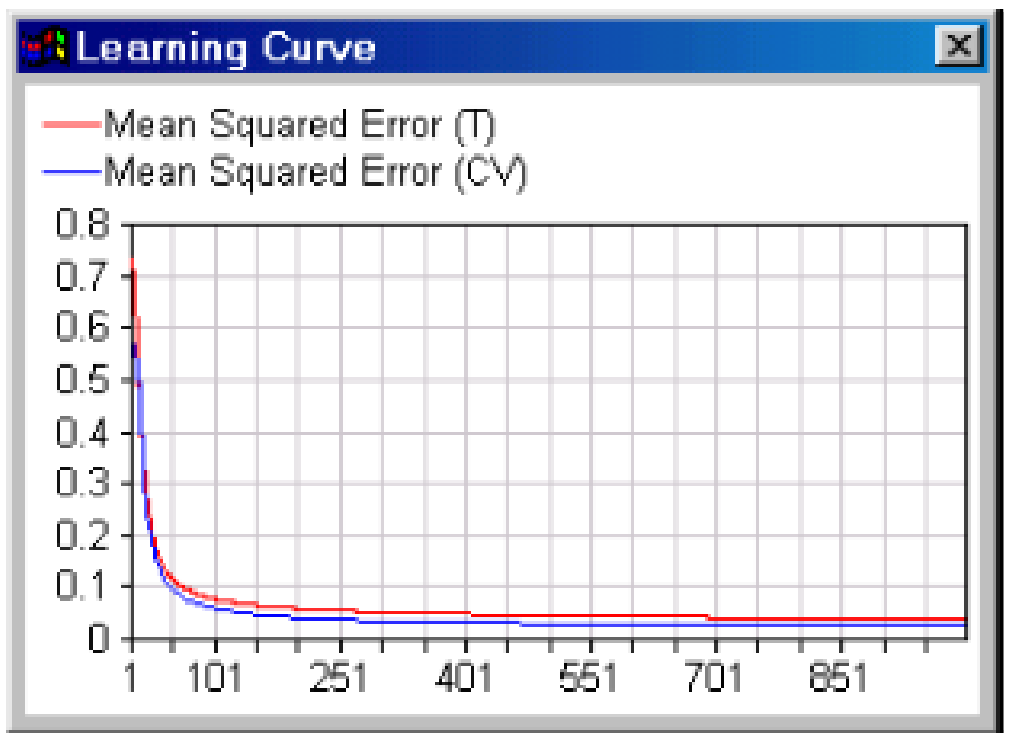

Figure 4: Sample of successful learning curve

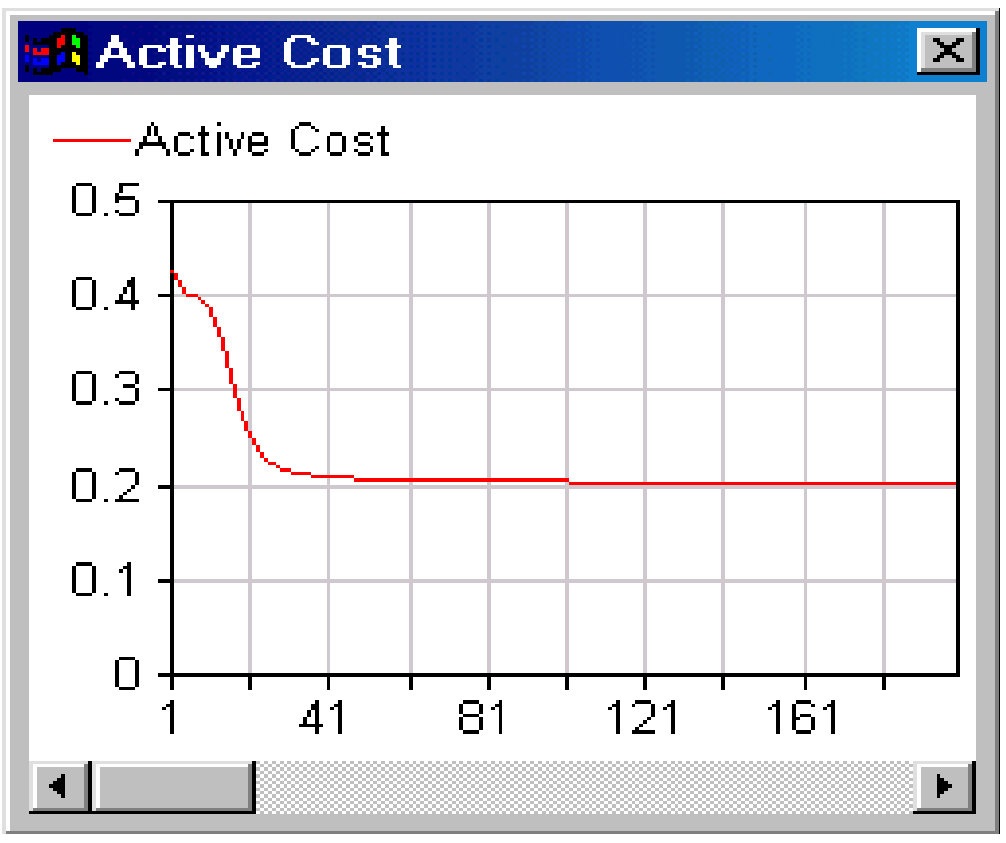

Figure 5: Sample of unsuccessful learning curve

### 2.7.2. THE CORRELATION COEFFICIENT (R)

The performance of the network output to the desired output can be measured with mean square error (MSE) value, but it doesn't necessarily tell the direction of movement of the two set of data, hence the need for the correlation coefficient. The correlation coefficient between a network output (g) and a desired output (f) is explained as the ratio of the covariance between the input and the desired output data over the product of their standard deviation. This is shown in equation 4.

$$R = \sum_{i=1}^{N} (Y_i - \bar{Y}_i)(X_i - \bar{X}_i) / \{ [\sum_{i=1}^{n} (Y_i - \bar{Y}_i)^2] [\sum_{i=1}^{n} (X_i - \bar{X}_i)^2] \}^{1/2} \quad (4)$$

The correlation coefficient ranges between (-1 and 1). When R =1 there is a perfect positive linear correlation between X and Y, which means they vary by the same amount. If R = -1, there is a perfectly linear negative correlation between X and Y, which means that they vary in opposite ways. When R = 0 there is no correlation between variables X and Y. Other intermediate values of R describe partial correlations. For instance, a value of R = 0.91 means that the fit of the model to the data is reasonably good for both data variables.

### 2.7.3. MEAN SQUARED ERROR (MSE)

The mean squared error was used in model design to determine how well the network output fits the desired output. It does not show whether the two sets of data move in the same direction. For instance, by simply scaling the network output, the mean square error can be changed without changing the directionality of the data. Also, the mean square error is simply two times the average cost as shown in equation 5.

$$MSE = \left( \sum_{j-0}^{p} \sum_{i-0}^{n} (d_{ij} - y_{ij})^2 \right) / (N * P) \quad (5)$$

Where, P = Number of output processing elements, N = Number of exemplars in the data set, $Y_{ij}$ = Network output for exemplars i at processing element j, $d_{ij}$ = Desired output for exemplar i at processing element j

### 2.7.3. NORMALIZED MEAN SQUARED ERROR (NMSE)

MSE is one of the major performance parameters and can be expressed as

$$NMSE = P (N * MSE) / \left( \sum_{j-0}^{p} N \sum_{i-0}^{N} d_{ij}^2 - \left(\sum_{i-0}^{N} d_{ij}\right)^2 \right) / N \quad (6)$$

Where, P = Number of output processing elements, N = Number of exemplars in the dataset, MSE = Mean Squared Error, $d_{ij}$ = Desired output of exemplar i at processing element j.
It is also used to determine how well the network output fits the desired output.

### 2.7.4 MEAN ABSOLUTE ERROR (MAE)

The Mean absolute error (MAE) is the measure of accuracy in a fitted time series value in a datasets, specifically trending. It usually expresses accuracy as a percentage and is defined by the formula:

$$M = 1/n \sum_{t-1}^{n} ( (A_t - F_t) / A_t) \quad (7)$$

Where, $A_t$ is the actual value and $F_t$ is the forecast value.

### 3.RESULTS AND DISCUSSION

This study involves basically "REGRESSION ALGORITHMS". The Neurosolution6 software comprise of different neural network models that has its individual applications uniquely. The network models that can be used in regression type study are:

- Multilayer Perceptron (MLP) Network
- Generalized Feed Forward Neural Network (GFNN)
- Radial Basis Function (RBF) Network
- Time - Lag Recurrent Network (TLRN)
- Recurrent Network (RN)

To perform the modelling of the data sets, each model was tested with the data sets for each sector. 60 % of the data sets for training, 25% for cross validation and 15% for testing the model. The five different network models mentioned above were used to test the datasets, but the data sets in this study is a time series problem. Out of the five major topologies considered, the Time - lag recurrent network and recurrent network were more applicable in a time series, although, the other topologies were also considered. Another importance of this two topologies is because the data sets contains information in its time structure, that is, how the data change with time (time series); and the other three regression topologies are purely static classifiers. Also, Time lagged recurrent network and recurrent network topologies are very good in nonlinear time series prediction. The hidden layers were varied from 0 to 4, making five runs for each model. The processing element for each of the model were 14, since we have a multi- variant data sets, in which each processing element has an effect on the other processing elements. The number of times for training also termed as the epoch were ranged from 900 to 1000 to have a stabilize model. The learning curve, Breadboards and performance measure were factors used to determine the best performing Artificial Neural Networks (ANNs) model. Table 3 shows the results of the performance measures of the neural networks topologies on the data sets as generated by the Neurosolution6 software. The best performing network was measured by the network with the minimum mean squared error (MSE) as possible. It was observed that the network models at 0-hidden layers recorded the highest minimum mean squared error (MSE) across the topologies (Table 3). These lead to an indication that the network did not fully learn the problem at the 0- hidden – layer settings and could also indicate that the research problem being solved is not linearly separable but a multi- variant dependent problem which confirms the variable type in the data set.

Setting the number of hidden layers to 1 resulted in another pattern of observation across the network models. The mean squared error (MSE) value changed considerably showing that the network might have been able to learn the pattern within the datasets. Although, the highest values of mean squared error (MSE) were observed at this point indicating the networks lack enough degree of freedom to solve, or learn the problem thereby making the error to stabilize at high values. Better trends were observed between the hidden layers of 2, 3, and 4 across the network topologies, although, there was a drift between the 3- hidden layers and the 4- hidden layers network models. The 2- hidden layers gave the least mean squared error (MSE) followed by the 3- hidden layer models but there were increased in the mean square error for the 4- hidden layers indicating more likely an over learning or cramming of the network.

The correlation coefficient (R) was used to ascertain if the network predicted output moves in the same direction as the desired output. The target considered for each model is to observe a value close to 1 as possible in order to generate a good model. The time lagged recurrent and recurrent networks gave better values for the correlation coefficient, 0.97 and 0.99, respectively; in comparison with multilayer perceptron (0.41), generalized feed forward (0.40), and radial basis function (0.45).

Optimization is one of the goals of applying computing techniques. The modelling time was also an indication of the maximum time it took to model the networks by each network topologies. The Multilayer Perceptron (MLP) network gave the best average time (39 seconds) while the Recurrent Network (RN) took more time (420 seconds) in modelling the network having the least optimization.

The performance of the 0- hidden layer network which is not too promising indicates that the input/output variables relationship among the data sets is non linear. Therefore, a linear model such as the 0- hidden layer network is not too appropriate for this study. The results also indicated that there was probably not too much need for more than three hidden layers as a result of the downward drift of performance across the network topologies as the hidden layers increases. Therefore, the models generated with 1, 2, 3 hidden layers are appropriate.

Intermediate network sizes produces the best results, this confirms the argument of Occam that "Any learning machine should be sufficiently large to solve the problem, but not larger" (Adefowoju, 2003). Considering the time taken to model the networks, Recurrent Network (RN) model and the Time lagged Recurrent Network (TLRN) both appeared as the best performing topologies, but the time taken by the Recurrent Network was far more than that taken for modelling by the Time Lagged Recurrent Network (TLRN). In terms of optimization and other factors been considered, the Time Lagged Network can be considered as the best performing Network Model generated by the Neurosolution6 software. Figures 6 to 11 gives examples of the sample models generated by the Time Lagged Network model.

The Time Lagged Network model which was the best performing Artificial Neural Network (ANNs) in this study was further tested with sample data sets from the Food Beverage and Tobacco sector, which were strange to the model. The results gave a trend accuracy of 92.1% which is an indication of how the predicted or desired pollution load moves with the actual pollution load as shown in Table 5. Furthermore, the percentage trend between the desired and the actual output was estimated. It gave an average of 86.5% of the prediction across the pollutants. In which the desired output and actual output represents the result generated by the Artificial Neural networks (ANNs) and Industrial pollution projection System (IPPS), respectively.

Table 3. Performance Measure Generated by Neurosolution6 Software

| PERFORMANCE MEASURE | TIME - LAG RECURRENT (TLR)<br>MEMORY = FOCUSED<br>FUNCTION = GAMMA AXON<br>DEPTH IN SAMPLES : 10<br>TRAJECTORY LENGTH : 10<br>TRANSFER FUNCTION :<br>SIGMOID AXON<br>LEARNING RULE: MOMENTUM<br>STEP SIZE = 0.1<br>MOMENTUM = 0.7, 0.9 | | | | |
|---|---|---|---|---|---|
| HIDDEN LAYERS | 0 | 1 | 2 | 3 | 4 |
| Modelling time (sec) | 39 | 41 | 44 | 49 | 52 |
| EPOCH | 1000 | 1000 | 1000 | 1000 | 1000 |
| MEAN SQUARED ERROR (MSE) | 0.056 | 0.019 | 0.017 | 0.080 | 0.098 |
| NORMALIZED MEAN SQUARED ERROR (NMSE) | 0.306 | 1.799 | 0.208 | 0.108 | 0.659 |
| MEAN AVERAGE ERROR (MAE) | 0.141 | 0.721 | 0.982 | 0.465 | 0.574 |
| MIN ABSOLUTE ERROR | 0.371 | 0.228 | 0.383 | 0.278 | 0.157 |
| MAX ABSOLUTE ERROR | 0.672 | 0.890 | 0.538 | 0.404 | 0.777 |
| LINEAR CORRELATION COEFFICIENT (R) | 0.672 | 0.9999 | 0.960 | 0.970 | 0.809 |

| PERFORMANCE MEASURE | RECURRENT NETWORK (RN)<br>INPUT LAYER = AXON<br>RECURRENCY = PARTIALLY FULLY RECURRENT<br>TRANSFER FUNCTION = SIGMOID -AXON<br>LEARNING RULE:MOMENTUM<br>STEP SIZE = 0.1<br>MOMENTUM = 0.7, 0.9 | | | | |
|---|---|---|---|---|---|
| HIDDEN LAYERS | 0 | 1 | 2 | 3 | 4 |
| Modelling time (sec) | 234 | 260 | 390 | 400 | 420 |
| EPOCH | 1000 | 1000 | 1000 | 1000 | 1000 |
| MEAN SQUARED ERROR (MSE) | 0.099 | 0.042 | 0.022 | 0.029 | 0.590 |
| NORMALIZED MEAN SQUARED ERROR (NMSE) | 0.140 | 0.249 | 0.218 | 0.301 | 0.499 |
| MEAN AVERAGE ERROR (MAE) | 0.657 | 0.930 | 0.180 | 0.423 | 0.466 |
| MIN ABSOLUTE ERROR | 0.561 | 0.208 | 0.232 | 0.230 | 0.464 |
| MAX ABSOLUTE ERROR | 0.810 | 0.592 | 0.436 | 0.560 | 0.870 |
| LINEAR CORRELATION COEFFICIENT (R) | 0.777 | 0.820 | 0.999 | 0.899 | 0.990 |

| PERFORMANCE MEASURE | MULTILAYER PERCEPTRON (MLP)<br>TRANSFER FUNCTION = SIGMOID AXON<br>LEARNING RULE : MOMENTUM<br>STEP SIZE = 1.0<br>MOMENTUM = 0.7, 0.9 | | | | |
|---|---|---|---|---|---|
| HIDDEN LAYERS | 0 | 1 | 2 | 3 | 4 |
| Modelling time (sec) | 30 | 32 | 20 | 36 | 39 |
| EPOCH | 1000 | 1000 | 1000 | 1000 | 1000 |
| MEAN SQUARED ERROR (MSE) | 1.903 | 0.988 | 0.760 | 0.934 | 0.870 |
| NORMALIZED MEAN SQUARED ERROR (NMSE) | 0.374 | 0.160 | 0.173 | 0.581 | 0.947 |
| MEAN AVERAGE ERROR (MAE) | 1.333 | 1.273 | 0.900 | 2.111 | 3.230 |
| MIN ABSOLUTE ERROR | 5.575 | 5.902 | 6.189 | 4.646 | 7.478 |
| MAX ABSOLUTE ERROR | 15.672 | 10.068 | 11.285 | 10.914 | 13.196 |
| LINEAR CORRELATION COEFFICIENT (R) | 0.101 | 0.276 | 0.278 | 0.410 | 0.010 |

| PERFORMANCE MEASURE | GENERALIZED FEED FORWARD NETWORK (GFFN)<br>TRANSFER FUUNCTION = SIGMOID AXON<br>LEARNING RULE : MOMENTUM<br>STEP SIZE = 1.0<br>MOMENTUM = 0.7, 0.9 | | | | |
|---|---|---|---|---|---|
| HIDDEN LAYERS | 0 | 1 | 2 | 3 | 4 |
| Modelling time (sec) | 40 | 60 | 30 | 120 | 130 |
| EPOCH | 1000 | 1000 | 1000 | 1000 | 1000 |
| MEAN SQUARED ERROR (MSE) | 0.998 | 0.120 | 0.110 | 0.478 | 0.870 |
| NORMALIZED MEAN SQUARED ERROR (NMSE) | 0.373 | 0.511 | 0.348 | 0.191 | 0.131 |
| MEAN AVERAGE ERROR (MAE) | 0.175 | 0.899 | 0.138 | 0.715 | 0.819 |
| MIN ABSOLUTE ERROR | 0.281 | 0.222 | 0.251 | 0.188 | 0.131 |
| MAX ABSOLUTE ERROR | 0.730 | 0.417 | 0.971 | 0.975 | 0.890 |
| LINEAR CORRELATION COEFFICIENT (R) | 0.481 | 0.275 | 0.449 | 0.404 | 0.51 |

| PERFORMANCE MEASURE | RADIAL BASIS FUNCTION (RBF)<br>CLUSTER CENTER = 80<br>COMPETITIVE RULE = CONSCIENCE FULL<br>METRIC = EUCLIDEAN<br>TRANSFER FUNCTION = SIGMOID AXON<br>LEARNING RULE = MOMENTUM<br>STEP SIZE 1.0<br>MOMENTUM 0.9,LEARNING RATE 0.01 TO 0.001 | | | | |
|---|---|---|---|---|---|
| HIDDEN LAYERS | 0 | 1 | 2 | 3 | 4 |
| Modelling time (sec) | 128 | 140 | 60 | 158 | 113 |
| EPOCH | 1000 | 1000 | 1000 | 1000 | 1000 |
| MEAN SQUARED ERROR (MSE) | 1.599 | 0.713 | 0.616 | 0.818 | 0.961 |
| NORMALIZED MEAN SQUARED ERROR (NMSE) | 0.873 | 0.717 | 0.27 | 0.480 | 0.980 |
| MEAN AVERAGE ERROR (MAE) | 0.105 | 0.673 | 0.754 | 0.28 | 0.557 |
| MIN ABSOLUTE ERROR | 0.259 | 0.344 | 0.328 | 0.222 | 0.252 |
| MAX ABSOLUTE ERROR | 0.412 | 0.591 | 0.590 | 0.898 | 0.780 |
| LINEAR CORRELATION COEFFICIENT (R) | 0.500 | 0.303 | 0.457 | 0.463 | 0.214 |

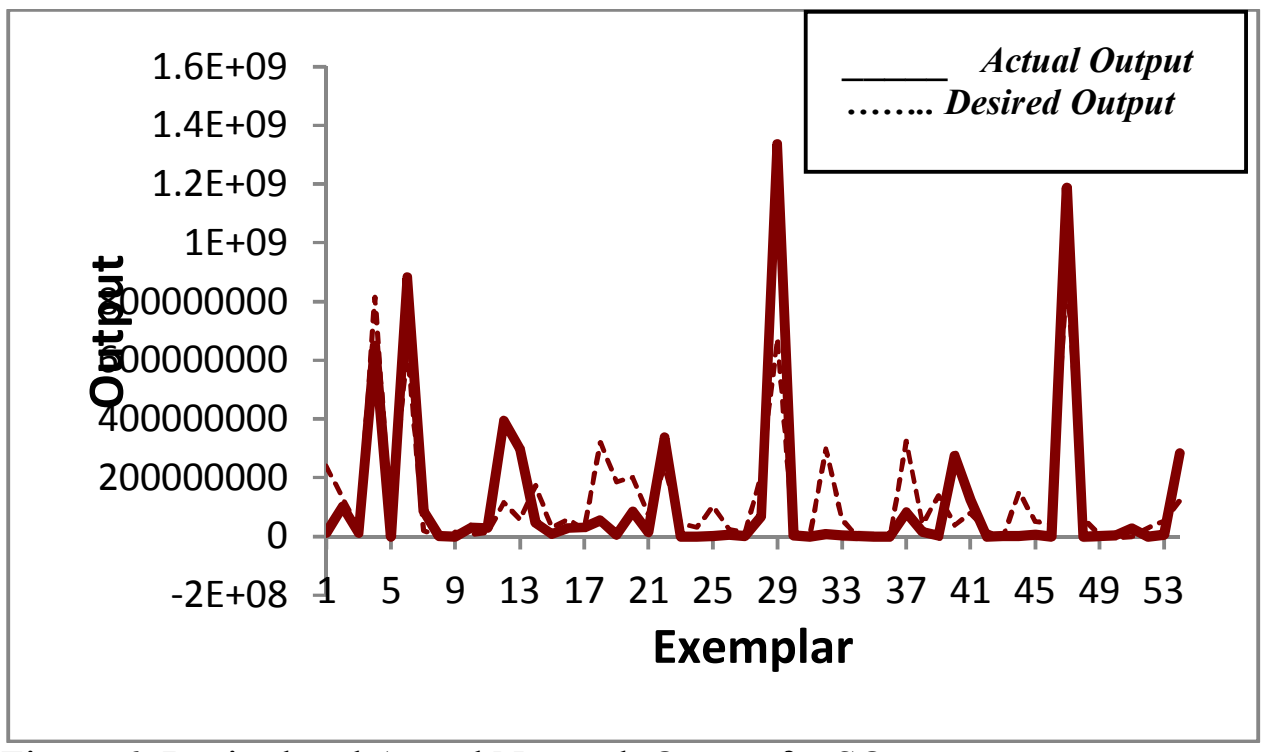

Figure 6: Desired and Actual Network Output for $SO_2$

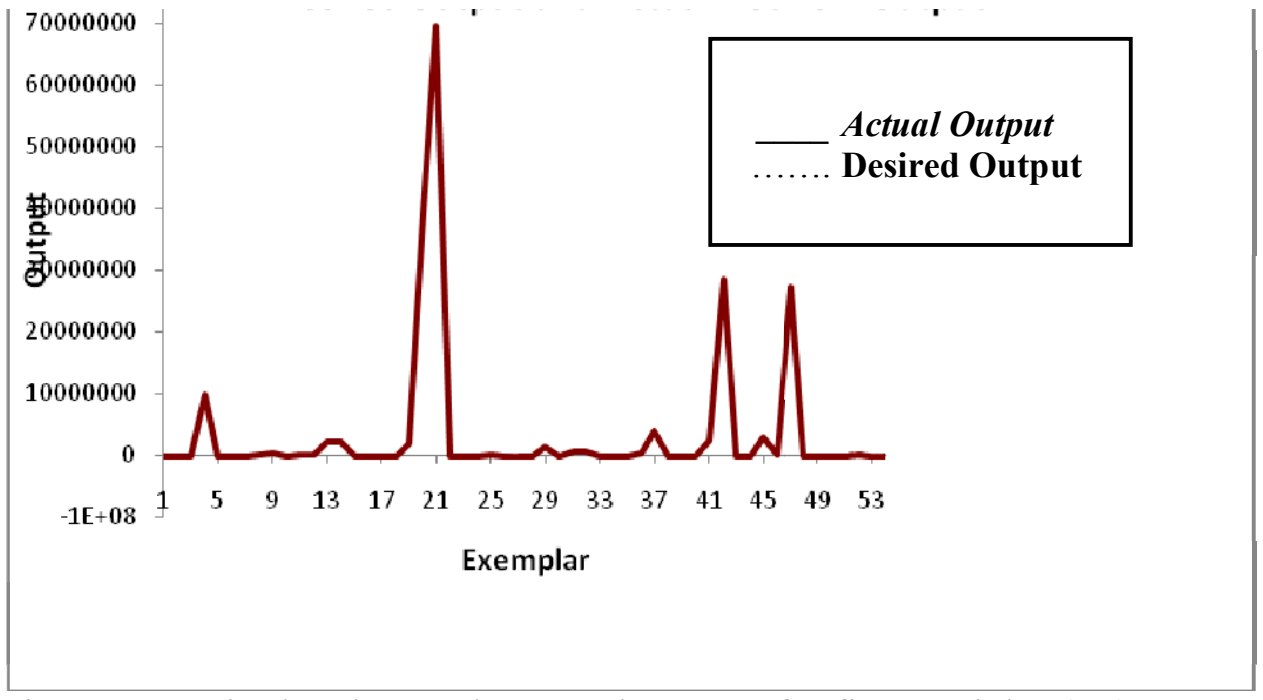

Figure 7: Desired and Actual Network Output for fine particles (FP)

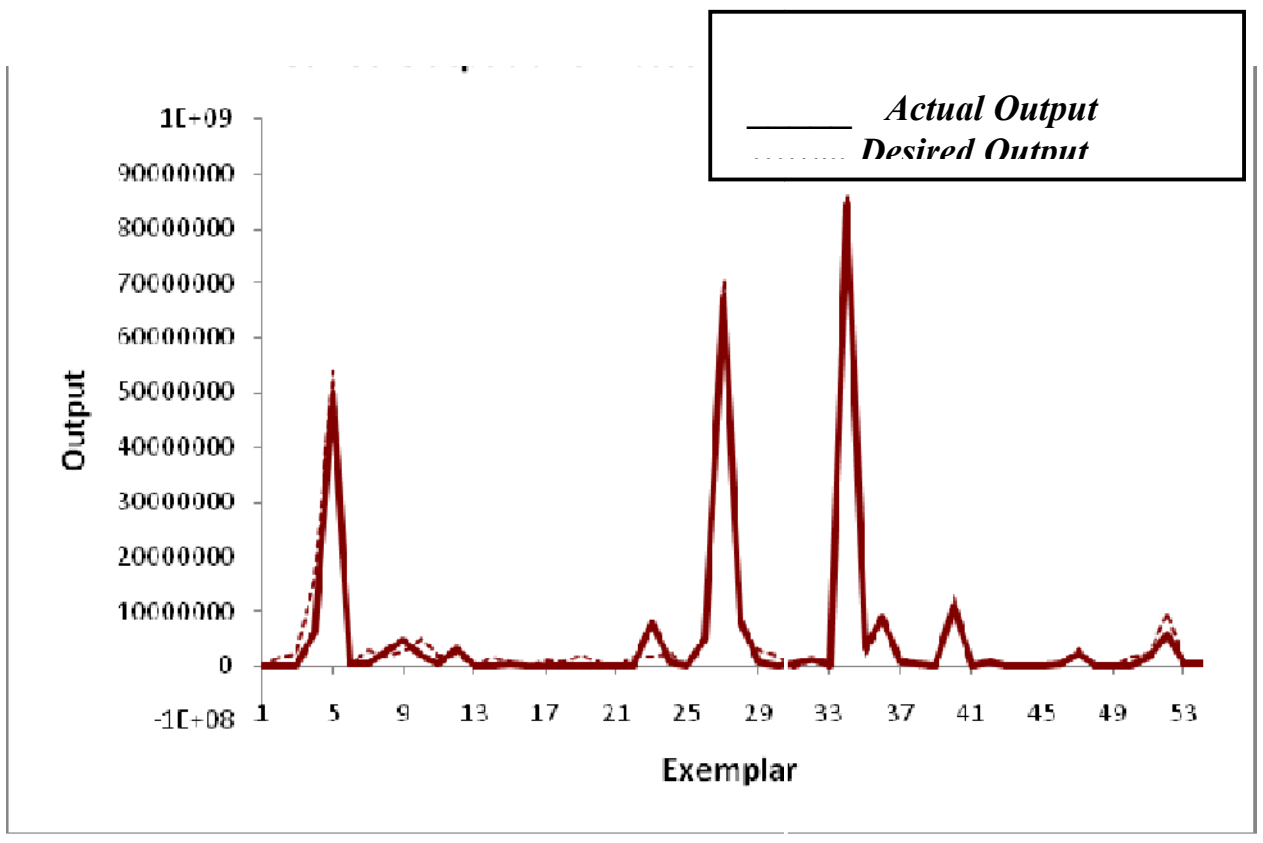

Figure 8: Desired and Actual Network Output for Total Suspended Particles (TSP)

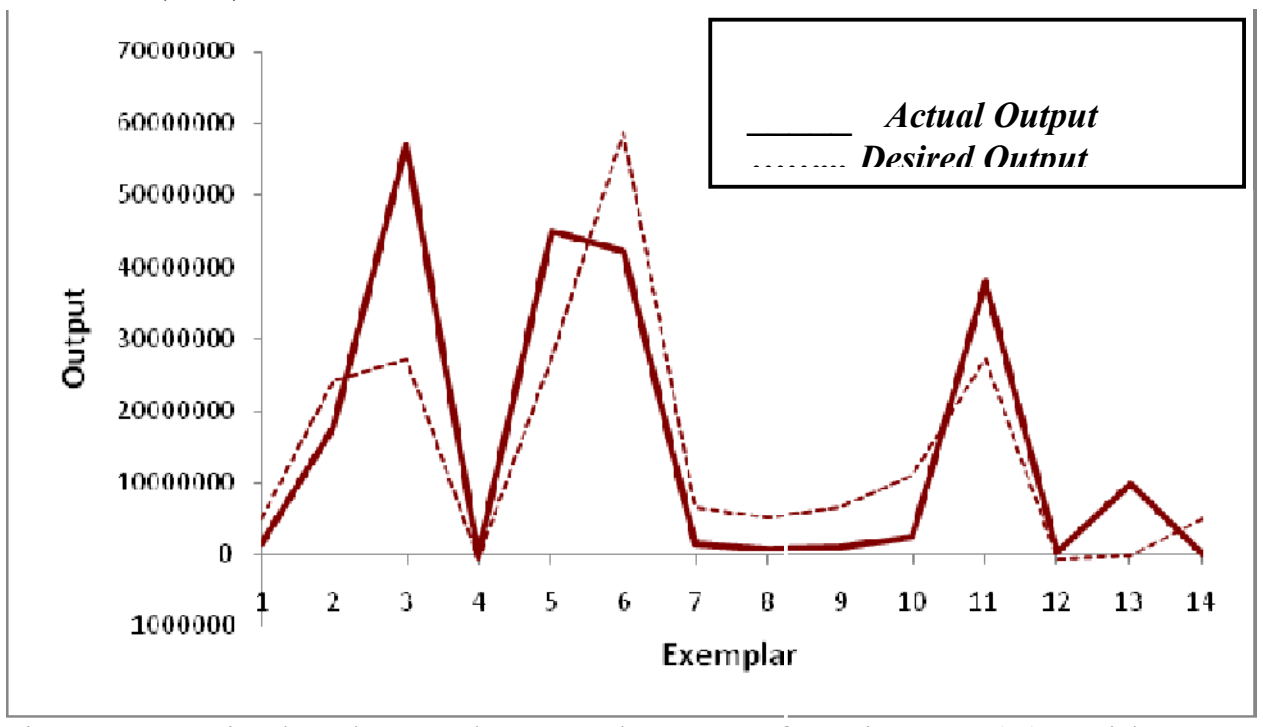

Figure 9: Desired and Actual Network Output for Nitrogen (II) Oxides ($NO_2$)

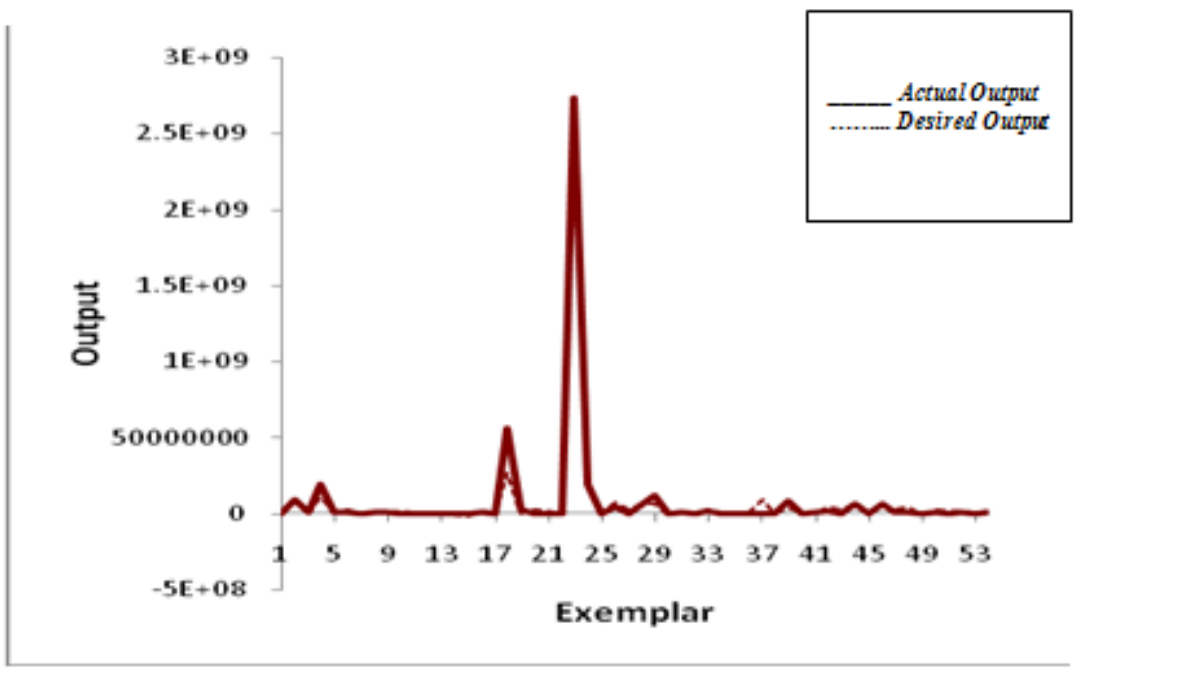

Figure 10: Desired and Actual Network Output for Toxic Chemical to land (TCLAND)

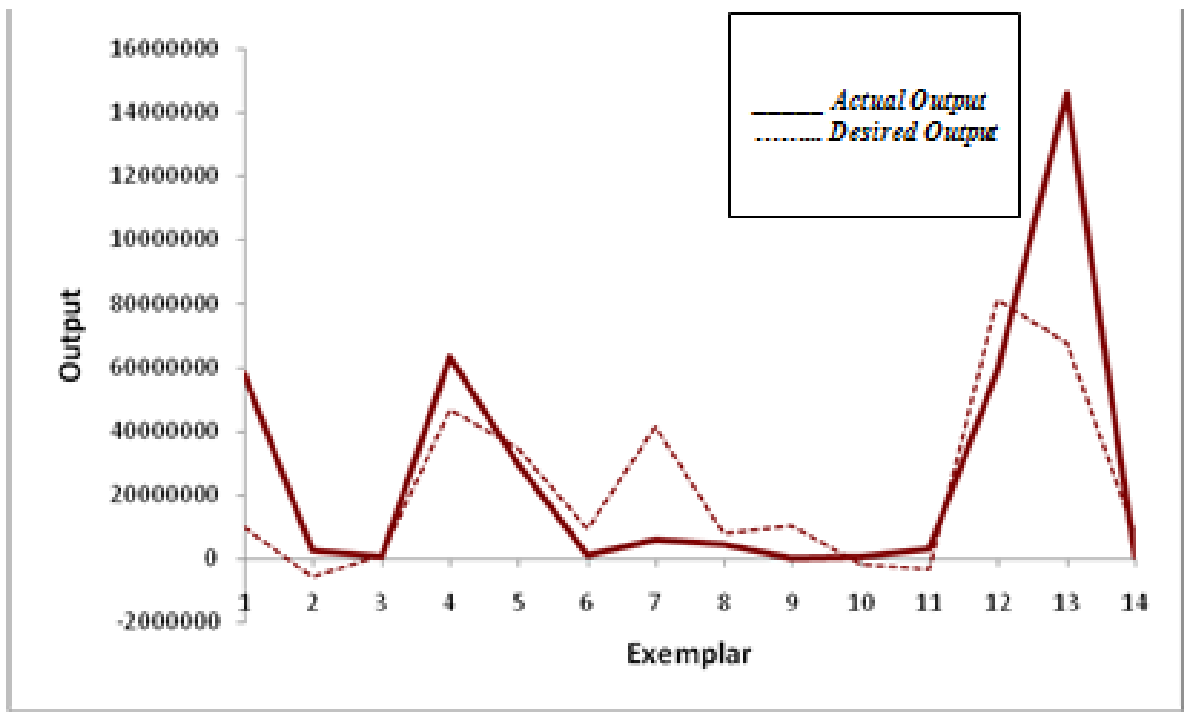

Figure 11: Desired and Actual Network Output for Toxic Chemical to water (TCWATER)

Table 6: Desired and the Actual Network Output with respect to employment for Food, Beverages & Tobacco sector (FBT) (tons)

| Pollutants | Desired output (tonnes) | Actual output (tonnes) |
|---|---|---|
| SULPHATE OXIDE($SO_2$) | 101.606 | 89.511 |
| NITROGEN (II) OXIDE ($NO_2$) | 237 | 242.140 |
| CARBON MONOXIDE (CO) | 169 | 134.241 |
| VOLATILE ORGANIC COMPOUND (VOC) | 118.994 | 133.881 |
| FINE PARTICLES (FP) | 165 | 123.831 |
| TOTAL SUSPENDED PARTICLES (TSP) | 235 | 218.958 |
| TOXIC CHEMICAL AIR (TC AIR) | 170 | 175.089 |
| TOXIC CHEMICAL LAND (TCLAND) | 165 | 93.197 |
| TOXIC CHEMICAL WATER (TCWATER) | 264 | 122.059 |
| TOXIC METAL (TMAIR) | 155 | 166.216 |
| TOXIC METAL LAND (TMLAND) | 78.237 | 69.140 |
| TOXIC METAL WATER (TMWATER) | 115 | 122.216 |
| BIOCHEMICAL OXYGEN DEMAND (BOD) | 106.352 | 117.425 |
| TOTAL SUSPENDED SOLIDS (TSS) | 130 | 116.707 |

## 4. CONCLUSION

The choice of the neural network topologies cannot be made without a run through their performance. In this study, five major network topologies were taken into considerations which are Radial basis, Generalized Feed Forward Networks (GFFN), Multilayer Perceptron (MLP),

Time Lagged Recurrent Networks (TLRN) and Recurrent Networks (RN).The hidden layers were varied between 0 to 4 for each network. The performance 2 and 3 hidden layers showed that the network learns better within the range of 2 and 3 hidden layers. The learning curves were also considered and the performance at 0 and 4 hidden layers showed that the network did not model the data sets at the setting of either 0 or 4. At first, the Multilayer Perceptron Network (MLP) which appeared to be well applicable in divers field of study (ecology) with a range of data sets in various study was considered as the best topology, but the data sets was not a static data sets, thus, it changed with time (time series). Time Lagged Recurrent Network (TLRN) and Recurrent Network (RN) were then considered. They both performed excellently well but optimization was a factor that was considered in picking the best between the two networks. The modelling time for the Recurrent Network topology was higher in comparison with the Time Lagged Recurrent Network (TLRN) model. The Time Lagged Recurrent Network (TLRN) model was considered as the best performing Artificial Neural Networks (ANN) model in this study considering all the factors such as modelling time, mean squared error, normalized mean squared error, mean average error, mean absolute error, max absolute error and linear correlation coefficient. Thus, Artificial intelligence (AI) is a diverse field that can be applied in environmental studies.

**First Author** Dr Adesesan. B Adeyemo is a lecturer at the Computer Science Department, University of Ibadan. His research interests include Data/Text mining, Networking and Internet Computing.

**Second Author** Dr Adebola A Oketola is a lecturer at the Department of Chemistry University of Ibadan she obtained her Doctoral degree in the year 2007 at the University of Ibadan, she is a member of Chemical society of Nigeria, Waste management society of Nigeria. Her research areas are Environmental modelling, persistent organic pollutant analysis, nanotechnology and chemical sensor.

**Third Author** Mr Emmanuel Olumuyiwa Adetula has a Masters degree in Computer Science from the University of Ibadan (2010); He is a Lecturer at the Federal University Lafia, Nigeria His research interest are Data mining and Artificial intelligence and its applications to other fields.

**Fourth Author** Professor O Osibanjo obtained his Doctoral degree in the year 1976 from the University of Birmingham and became a professor in the year 1989. He lectures in the Department of Chemistry University of Ibadan and the Director of Basel convention coordinating centre for the African region. Research interests are Environmental modelling, persistent organic pollutant analysis and e- waste.